%% file: main.tex
\documentclass{bvm} 

\usepackage{array}
\usepackage{tabularray}
\usepackage{adjustbox}
\usepackage{acronym}
\usepackage{lipsum}
\acrodef{SAM}{Segment Anything Model}
\acrodef{cmri}[cMRI]{cardiac MRI}
\acrodef{hd}[HD]{Hausdorff Distance}
\acrodef{mad}[MAD]{Mean Absolute Distance}
\acrodef{lv}[LV]{Left Ventricle}
\acrodef{rv}[RV]{Right Ventricle}

\newcolumntype{P}[1]{>{\centering\arraybackslash}p{#1}}

\begin{document}

\newcommand{\bvmyear}{2023}

\selectlanguage{english} 

\title{Influence of Prompting Strategies on Segment Anything Model (SAM) for Short-axis Cardiac MRI segmentation}
\titlerunning{Influence of Prompting Strategies on SAM}


\author{
         Josh \lname{Stein} \inst{1,2}, 
	Maxime \lname{Di Folco} \inst{1}, 
	Julia A. \lname{Schnabel} \inst{1,2,3}, 
}
\authorrunning{J. Stein et al.}

%

\institute{
\inst{1}  Institute of Machine Learning in Biomedical Imaging, Helmholtz Munich, Neuherberg, Germany\\
\inst{2} Technical University of Munich, Munich, Germany\\
\inst{3}  King’s College London, London, UK
}

\email{maxime.difolco@helmholtz-munich.de}

\maketitle

\begin{abstract}

The Segment Anything Model (SAM) has recently emerged as a significant breakthrough in foundation models, demonstrating remarkable zero-shot performance in object segmentation tasks. While SAM is designed for generalization, it exhibits limitations in handling specific medical imaging tasks that require fine-structure segmentation or precise boundaries. In this paper, we focus on the task of cardiac magnetic resonance imaging (cMRI) short-axis view segmentation using the SAM foundation model. We conduct a comprehensive investigation of the impact of different prompting strategies (including bounding boxes, positive points, negative points, and their combinations) on segmentation performance. We evaluate on two public datasets using the baseline model and models fine-tuned with varying amounts of annotated data, ranging from a limited number of volumes to a fully annotated dataset. Our findings indicate that prompting strategies significantly influence segmentation performance. Combining positive points with either bounding boxes or negative points shows substantial benefits, but little to no benefit when combined simultaneously. We further observe that fine-tuning SAM with a few annotated volumes improves segmentation performance when properly prompted. Specifically, fine-tuning with bounding boxes has a positive impact, while fine-tuning without bounding boxes leads to worse results compared to baseline.

\end{abstract}

\section{Introduction}

The \ac{SAM} \cite{Kirillov_2023} represents a notable and recent breakthrough in foundation models for computer vision applications. \ac{SAM} is a model pre-trained on a dataset containing more than 1 billion masks from 11 million images. It  demonstrates an exceptional zero-shot performance, often rivalling or out-performing prior fully supervised models. This model is specifically engineered for object segmentation with diverse prompting strategies in the form of points, bounding boxes, and/or textual input. SAM is designed for a broad generalisation and depth across a variety of segmentation tasks. Nonetheless, in medical imaging we can encounter more specific tasks such as segmentation of fine-structure (e.g. lung nodules) or the need of clear boundaries of segmentation mask (e.g. tumor region), where \ac{SAM} network is limited \cite{Kirillov_2023}. Many recent works have adapted \ac{SAM} and explored different prompting strategies for medical segmentation tasks and compared the performance to state-of-the-art segmentation networks on a wide range of datasets, i.e. different imaging modalities and organs \cite{Huang_2023, Cheng_2023, He_2023, Ma_2023}. Specially, Ma et al. \cite{Ma_2023} introduced MedSAM which outperforms a baseline U-Net by fine-tuning the image encoder and mask decoder on an extensive medical image dataset composed of more than 1 million images. However, the impact of prompting strategies for specific tasks has not been thoroughly investigated. 

In this paper, we aim to investigate the role of the SAM foundation model for the specific task of cMRI short-axis view segmentation. We assess the impact of different prompting strategies (bounding boxes, positive points, negative points and their combinations) on segmentation performance using two publicly available datasets. We further investigate the effect of fine-tuning the model on a varying number of annotated data: from few volumes to a fully annotated dataset.

\section{Methods}

\subsection{Segment Anything Model (SAM):} 

\ac{SAM} \cite{Kirillov_2023} is the largest foundation model for image segmentation. The model is composed of three parts: an image encoder, a prompt encoder, and a mask decoder. The model can be used without any prompt information, in which case the model segments as many objects as possible. Alternatively, the model can be prompted using text inputs, bounding boxes and positive or negative points. These prompts allow the model to `focus' on regions of interest. In this paper, we experiment with different prompting strategies using combinations of bounding boxes, positive points and negative points with or without fine-tuning \ac{SAM}. We fine-tune on a varying number of cases, from 8 to the entire dataset, and learn only the mask decoder (i.e. we freeze the image encoder and prompt encoder, and propagate gradients only through the mask decoder). The default ViT-H model is used for all inference and fine-tuning experiments. 
We run inference on \ac{SAM} per class (i.e. we use the \ac{lv}, the myocardium, and the \ac{rv} independently for our tasks) and combine the results into a single output segmentation. We refer to the \textit{baseline} model when referencing experiments without fine-tuning in the experimental section. We evaluate model performance using Dice score, \ac{hd} and the \ac{mad}. Models were fine-tuned for 100 epochs. We use the Adam optimiser ($\beta_1$ = 0.9, $\beta_2$ = 0.999).
An initial learning rate of 1e-4 was empirically determined. We reduce the learning rate
on epoch loss plateau with a patience of 5 by a factor of 0.1. A batch size of 1 was used
due to GPU memory limitations

\section{Experiments and results}

\begin{table}[t!]
\centering

\input{tables/ACDC_baseline}
\caption{Baseline SAM inference results on the \textbf{ACDC} dataset. Positive and negative sample counts are per each segmentation class. All Dice standard deviations are less than 0.15. Unless shown, HD standard deviations are less than 5mm and MAD standard deviations are less than  2mm.}
\label{tab:baseline_sam_acdc}
\end{table}

\subsection{Dataset}

We experiment with two well-established cardiac imaging datasets, sourced from the the Automatic Cardiac Diagnosis Challenge (ACDC) \cite{bernard2018deep} and the Multi-Centre, Multi-Vendor, and Multi-Disease Cardiac Segmentation Challenge (M\&Ms) \cite{campello2021multi}. These datasets exhibit inherent sparsity, as they only provide annotations at end-diastolic and end-systolic phases. The ACDC dataset contains 100 training cases (providing a total of 200 annotated volumes - one for each cardiac phase). 160 volumes were reserved for training and the remaining 40 for validation. The test dataset is composed of 50 cases. Similarly, the M\&Ms dataset is composed of 150 training cases (corresponding to 300 training volumes). We split this data into 240 training volumes and 60 validation volumes - the test set contains 136 cases. Notably, the M\&Ms dataset encompasses images from diverse vendors and centers. We use the dataset split described in the original work by Campello et al. \cite{campello2021multi}.

We use a variety of pre-processing techniques inspired by nnU-Net \cite{isensee2021nnu}. These include random scaling, rotations, Gaussian blurring, brightness, contrast, Gamma correction and mirroring. Images were resized to 224 x 224. 

\subsection{Baseline model}

\begin{table}[hbt!]
\centering

\input{tables/MnMs_baseline}
\caption{Baseline SAM inference results on the \textbf{M\&Ms} dataset. Positive and negative sample counts are per each segmentation class. All Dice standard deviations are less than 0.15. Unless shown, HD standard deviations are less than 5mm and MAD standard deviations are less than 2mm.}
\label{tab:baseline_sam_mnms}
\end{table}

We firstly experimented using the \ac{SAM} \textit{baseline} model to determine the best prompt setup. We ran inference using combinations of positively sampled points, negatively sampled points and presence/absence of bounding boxes. The Dice score, \ac{hd} and \ac{mad} are reported in Tables  \ref{tab:baseline_sam_acdc} and \ref{tab:baseline_sam_mnms} for the ACDC and M\&Ms datasets respectively. We observe that in all configurations (with or without bounding boxes and for any number of positive points) using negative samples improves performance for both datasets. The change in improvement reduces when bounding boxes are included in the prompt. For example, on the ACDC dataset (Table. \ref{tab:baseline_sam_acdc}) with 2 positive samples (and a negative sample) the Dice score improves from 0.48 to 0.66 without a bounding box and from 0.65 to 0.69 with a bounding box.

\begin{figure}[t!]
    \centering
    \includegraphics[width=0.5\textwidth]{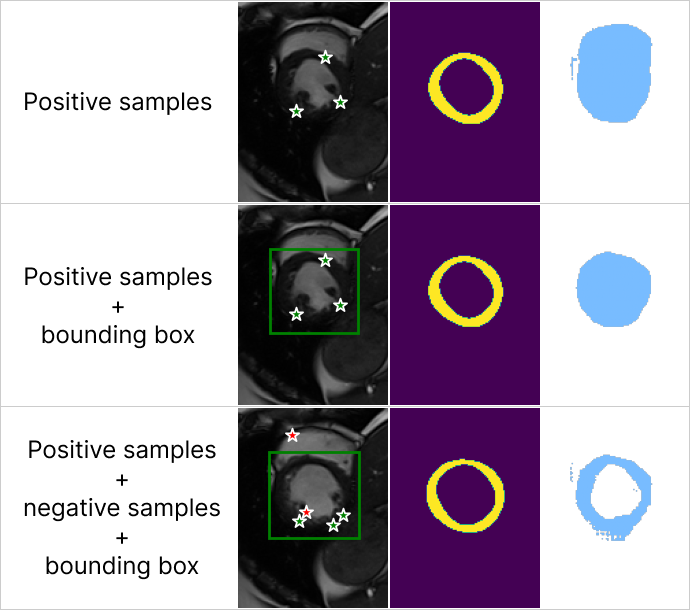}
    \caption{Effect of different prompting strategies when running inference on the myocardium.}
    \label{fig:myocardium_prompting}
\end{figure}

The use of either negative sample points or bounding boxes provide rich spatial information, allowing the network to differentiate between the \ac{lv} and the myocardium (which envelops the \ac{lv}). Specifically for the myocarduim, \ac{SAM} generates poor segmentation masks when prompting with only positive points, as shown in Figure. \ref{fig:myocardium_prompting}.

Using an increased number of positive sample points tends to have an effect only when prompting with neither bounding boxes nor negative points (the Dice score improves from 0.48 to 0.60 between 2 and 5 positive points on the ACDC dataset and from 0.46 to 0.57 on the M\&Ms dataset). 

\subsection{Fine-tuned model}


We fine-tune using prompts that include two positive points and one negative point, with or without bounding boxes. The results of fine-tuning with bounding boxes are presented in Table. \ref{tab:sam_finetuned_reduced_data}. We observe that fine-tuning improves performance on both datasets, with a larger improvement of 15\% achieved on the ACDC dataset. We further observe that training with a greater number of volumes generally increases performance, although the gains are limited. This is especially true for the M\&Ms dataset, where fine-tuning with 8 volumes yields a similar Dice score compared to fine-tuning with the fully annotated dataset. Compared to nnU-Net models trained from scratch, we find that nnU-Net models are able to out-perform fine-tuned SAM models on both datasets \cite{stein2023sparse}. That said, on the ACDC dataset the SAM models tend to perform better when using less than 32 training volumes. Table. \ref{tab:finetune_no_bbox}  presents the results when fine-tuning without bounding boxes. We observe that fine-tuning without bounding boxes leads to significantly worse performance on both datasets. From our earlier inference results, we know that not using bounding boxes tends to give very large surface distances. It is possible that the produced segmentations are too large, and the model is not able to learn well-differentiated borders.

\begin{table}[ht!]
    \centering
    \input{tables/fine_tuned_with_bbox}

    \caption{Inference results for SAM models fine-tuned with limited training data. The models were prompted \textbf{with bounding boxes}, two positive sample points and one negative sample per class. Dice standard deviations, HD standard deviations and MAD standard deviations are less than 0.1, 3mm and 1.2mm for all models respectively. Note that the ACDC dataset only has 160 volumes.}
    \label{tab:sam_finetuned_reduced_data}
\end{table}

\begin{table}[ht!]
    \centering
    \input{tables/fine_tuned_without_bbox}
    \caption{Inference results of fine-tuned SAM models. The models were prompted \textbf{without bounding boxes}, two positive sample points and one negative sample per class. Dice standard deviations are less than 0.1 for all models. Unless shown, MAD standard deviations are less than 2mm.}
    \label{tab:finetune_no_bbox}
\end{table}

\section{Discussion and conclusion}

In this paper, we experimented with \ac{SAM} foundation models for \ac{cmri} short-axis view segmentation. Specifically, we evaluated different prompting strategies, using bounding boxes, positive and negative points in order to determine their influence on segmentation performance. We initially experiment with the \textit{baseline} model and show that the prompting strategies have a large influence on performance. Our analyses demonstrate that combining positive points either with bounding boxes or negative points has a significant impact, but there is little to no difference when combining bounding boxes and negative points simultaneously. In our experiments, we re-use positive samples for particular classes as negative samples for other classes. Therefore, there is no additional cost of sampling when using negative samples. Models inferred with negative samples yield results very similar to using bounding boxes. For baseline segmentation, we recommend using more positively sampled points (which allows for greater re-use as negative samples), rather than using bounding boxes.

We further fine-tuned \ac{SAM} using a varying number of annotated volumes. We show that fine-tuning with a limited number of volumes can enhance segmentation performance when appropriately prompted. For this task, we observed that fine-tuning with bounding boxes positively impacts performance, while the absence of bounding boxes resulted in worse results compared to baseline. This is surprising, considering that baseline inference yielded comparable performance between models prompted with and without bounding boxes. We observe that fine-tuned SAM models can be outperformed by specialized nnU-Net models, even with the same amount of training data. Finally, we note that the cost of fine-tuning is relatively small, but can yield significant improvements. When pre-computing image encodings, fine-tuning is fast and efficient. 

\printbibliography

\end{document}

%% file: tables/ACDC_baseline.tex
\begin{adjustbox}{max width=\textwidth}
\begin{tabular}{|c|c|c|P{1cm}|c|c|} 
\hline
Bounding boxes     & Pos. samples       & Neg. samples & Dice          & HD (mm)           & MAD(mm)           \\ 
\hline
\multirow{2}{*}{N} & \multirow{2}{*}{2} & 0            & 0.48          & 38.29 $\pm$ 14.14 & 12.34 $\pm$ 5.06  \\ 
\cline{3-6}
                   &                    & 1            & 0.66          & 13.73 $\pm$ 8.31  & 4.67 $\pm$ 2.92    \\ 
\hline
\multirow{2}{*}{N} & \multirow{2}{*}{3} & 0            & 0.53          & 34.09 $\pm$ 13.06 & 10.93 $\pm$ 4.36  \\ 
\cline{3-6}
                   &                    & 1            & 0.68          & 13.65 $\pm$~8.10  & 4.61 $\pm$ 2.79   \\ 
\hline
\multirow{2}{*}{N} & \multirow{2}{*}{5} & 0            & 0.60          & 25.00 $\pm$ 12.01 & 8.24 $\pm$ 3.86   \\ 
\cline{3-6}
                   &                    & 1            & \textbf{0.69} & 13.33 $\pm$ 7.51  & 4.53 $\pm$ 2.51   \\ 
\hline
\multirow{2}{*}{Y} & \multirow{2}{*}{2} & 0            & 0.65          & 14.19             & 5.31              \\ 
\cline{3-6}
                   &                    & 1            & \textbf{0.69} & \textbf{12.21}    & \textbf{4.14}     \\ 
\hline
\multirow{2}{*}{Y} & \multirow{2}{*}{3} & 0            & 0.65          & 14.21             & 5.35              \\ 
\cline{3-6}
                   &                    & 1            & 0.68          & 12.79             & 4.39              \\ 
\hline
\multirow{2}{*}{Y} & \multirow{2}{*}{5} & 0            & 0.66          & 14.03             & 5.28              \\ 
\cline{3-6}
                   &                    & 1            & 0.68          & 13.45             & 4.65              \\
\hline
\end{tabular}
\end{adjustbox}

%% file: tables/MnMs_baseline.tex
\begin{adjustbox}{max width=\textwidth}
\begin{tabular}{|c|c|c|P{1cm}|c|c|} 
\hline
Bounding boxes     & Pos. samples       & Neg. samples & Dice          & HD (mm)           & MAD(mm)           \\ 
\hline
\multirow{2}{*}{N} & \multirow{2}{*}{2} & 0            & 0.46          & 41.30 $\pm$ 14.65 & 13.55 $\pm$ 5.36  \\ 
\cline{3-6}
                   &                    & 1            & 0.65          & 14.65 $\pm$ 8.75  & 4.97 $\pm$ 3.09   \\ 
\hline
\multirow{2}{*}{N} & \multirow{2}{*}{3} & 0            & 0.48          & 37.66 $\pm$ 13.75 & 12.29 $\pm$ 4.74  \\ 
\cline{3-6}
                   &                    & 1            & 0.66          & 14.24 $\pm$ 8.44  & 4.91 $\pm$ 2.96   \\ 
\hline
\multirow{2}{*}{N} & \multirow{2}{*}{5} & 0            & 0.57          & 27.63 $\pm$ 12.20 & 9.12 $\pm$ 3.99   \\ 
\cline{3-6}
                   &                    & 1            & 0.67          & 14.09 $\pm$ 8.09  & 4.83 $\pm$ 2.73   \\ 
\hline
\multirow{2}{*}{Y} & \multirow{2}{*}{2} & 0            & 0.66          & 12.90             & 4.94              \\ 
\cline{3-6}
                   &                    & 1            & \textbf{0.69} & \textbf{11.77}             & \textbf{4.05}     \\ 
\hline
\multirow{2}{*}{Y} & \multirow{2}{*}{3} & 0            & 0.66          & 12.89             & 4.95              \\ 
\cline{3-6}
                   &                    & 1            & 0.69          & 11.96             & 4.17              \\ 
\hline
\multirow{2}{*}{Y} & \multirow{2}{*}{5} & 0            & 0.66          & 12.80             & 4.90              \\ 
\cline{3-6}
                   &                    & 1            & 0.68          & 12.39             & 4.36              \\
\hline
\end{tabular}
\end{adjustbox}

%% file: tables/fine_tuned_with_bbox.tex
\begin{adjustbox}{max width=\textwidth}
\begin{tblr}{
  cells = {c},
  cell{1}{3} = {c=9}{},
  cell{3}{1} = {r=3}{},
  cell{6}{1} = {r=3}{},
  vline{1,3,12} = {1}{},
  vline{1-12} = {2-8}{},
  hline{1-3,6,9} = {-}{},
}
        &          & Number of fine-tuning cases &      &      &      &      &               &               &      &               \\
Dataset & Metric   & 0 (baseline)                & 8    & 24   & 32   & 48   & 80            & 160           & 192  & 240           \\
ACDC    & Dice     & 0.69                        & 0.76 & 0.81 & 0.82 & 0.82 & 0.83          & \textbf{0.84} & -    & -             \\
        & HD (mm)  & 12.21                       & 9.92 & 7.42 & 7.06 & 7.65 & 6.33          & \textbf{5.78} & -    & -             \\
        & MAD (mm) & 4.14                        & 3.50 & 2.45 & 2.38 & 2.48 & 2.12 & \textbf{2.02} & -    & -             \\
M\&Ms    & Dice     & 0.69                        & 0.76 & 0.74 & 0.74 & 0.75 & 0.76          & 0.76          & 0.76 & \textbf{0.76} \\
        & HD (mm)  & 11.76                       & 8.12 & 7.30 & 7.21 & 7.07 & 7.82          & 6.37          & 11.77 & \textbf{6.05} \\
        & MAD (mm) & 4.05                        & 2.70 & 2.66 & 2.67 & 2.61 & 2.68          & 2.44          & 2.47 & \textbf{2.39} 
\end{tblr}
\end{adjustbox}

%% file: tables/fine_tuned_without_bbox.tex
\begin{adjustbox}{max width=\textwidth}
\begin{tblr}{
  cells = {c},
  cell{1}{3} = {c=7}{},
  cell{3}{1} = {r=3}{},
  cell{6}{1} = {r=3}{},
  vline{1,3-4,10} = {1}{},
  vline{1-10} = {2-8}{},
  hline{1-3,6,9} = {-}{},
}
        &          & Number of fine-tuning cases &                  &                  &                  &                  &                  &            \\
Dataset & Metric   & 0 (baseline)                & 8                & 24               & 48               & 80               & 160              & 240        \\
ACDC    & Dice     & \textbf{0.66}               & 0.61             & 0.57             & 0.58             & 0.59             & 0.56             & -          \\
        & HD (mm)  & \textbf{13.73 $\pm$ 8.31}   & 17.11 $\pm$ 6.10 & 20.37 $\pm$ 7.64 & 19.05 $\pm$ 6.93 & 19.56 $\pm$ 8.31 & 19.30 $\pm$ 6.03 & -          \\
        & MAD (mm) & \textbf{4.67 $\pm$ 2.92}    & 5.99             & 7.13 $\pm$ 2.60    & 6.52 $\pm$ 2.15  & 6.83 $\pm$ 2.57 & 7.02           & -          \\
M\&Ms     & Dice     & \textbf{0.65}                       & 0.54             & 0.62             & 0.57             & 0.58             & 0.58             & 0.59       \\
        & HD (mm)  & \textbf{14.65 $\pm$ 8.75}                  & 23.91 $\pm$ 8.49       & 15.71 $\pm$ 5.69       & 18.05 $\pm$ 5.32       & 17.75 $\pm$ 5.35       & 17.89 $\pm$ 5.70       & 16.71 $\pm$ 6.04 \\
        & MAD (mm) & \textbf{4.97 $\pm$ 3.09}                       & 8.73 $\pm$ 2.81       & 5.51             & 6.75             & 6.63            & 6.67 $\pm$ 2.08            & 6.10 $\pm$ 2.08       
\end{tblr}
\end{adjustbox}

%% file: ref.bib
@article{bernard2018deep,
  title={{Deep learning techniques for automatic MRI cardiac multi-structures segmentation and diagnosis: is the problem solved?}},
  author={Bernard, Olivier and Lalande, Alain and Zotti, Clement and others},
  journal={IEEE TMI},
  year={2018},
}

@article{campello2021multi,
  title={{Multi-Centre, Multi-Vendor and Multi-Disease Cardiac Segmentation: the M\&Ms Challenge}},
  author={Campello, Victor M and Gkontra, Polyxeni and Izquierdo, Cristian and others},
  journal={IEEE TMI},
  volume={40},
  number={12},
  pages={3543--3554},
  year={2021},
  publisher={IEEE}
}

@article{isensee2021nnu,
  title={{nnU-Net: a self-configuring method for deep learning-based biomedical image segmentation}},
  author={Isensee, Fabian and Jaeger, Paul F and Kohl, Simon AA and Petersen, Jens and Maier-Hein, Klaus H},
  journal={Nature methods},
  volume={18},
  number={2},
  pages={203--211},
  year={2021},
  publisher={Nature Publishing Group US New York}
}

@article{Kirillov_2023, title={Segment Anything}, DOI={10.48550/arXiv.2304.02643}, note={arXiv:2304.02643 [cs]}, number={arXiv:2304.02643}, publisher={arXiv}, author={Kirillov, Alexander and Mintun, Eric and Ravi, Nikhila and Mao, Hanzi and Rolland, Chloe and Gustafson, Laura and Xiao, Tete and Whitehead, Spencer and Berg, Alexander C. and Lo, Wan-Yen and Dollár, Piotr and Girshick, Ross}, year={2023} }

@article{Huang_2023, title={{Segment Anything Model for Medical Images?}}, url={http://arxiv.org/abs/2304.14660}, note={arXiv:2304.14660 [cs, eess]}, number={arXiv:2304.14660}, publisher={arXiv}, author={Huang, Yuhao and Yang, Xin and Liu, Lian and Zhou, Han and Chang, Ao and Zhou, Xinrui and Chen, Rusi and Yu, Junxuan and Chen, Jiongquan and Chen, Chaoyu and Chi, Haozhe and Hu, Xindi and Fan, Deng-Ping and Dong, Fajin and Ni, Dong}, year={2023} }

@article{Ma_2023, title={{Segment Anything in Medical Images}}, url={http://arxiv.org/abs/2304.12306}, note={arXiv:2304.12306 [cs, eess]}, number={arXiv:2304.12306}, publisher={arXiv}, author={Ma, Jun and He, Yuting and Li, Feifei and Han, Lin and You, Chenyu and Wang, Bo}, year={2023} }

@article{He_2023, title={{Computer-Vision Benchmark Segment-Anything Model (SAM) in Medical Images: Accuracy in 12 Datasets}}, url={http://arxiv.org/abs/2304.09324}, publisher={arXiv}, author={He, Sheng and Bao, Rina and Li, Jingpeng and others}, year={2023} }

@article{Cheng_2023, title={{SAM on Medical Images: A Comprehensive Study on Three Prompt Modes}}, url={http://arxiv.org/abs/2305.00035}, note={arXiv:2305.00035 [cs]}, number={arXiv:2305.00035}, publisher={arXiv}, author={Cheng, Dongjie and Qin, Ziyuan and Jiang, Zekun and Zhang, Shaoting and Lao, Qicheng and Li, Kang}, year={2023} }

@misc{stein2023sparse,
      title={Sparse annotation strategies for segmentation of short axis cardiac MRI}, 
      author={Josh Stein and Maxime {Di Folco} and Julia Schnabel},
      year={2023},
      eprint={2307.12619},
      archivePrefix={arXiv},
      primaryClass={eess.IV}
}
